\documentclass[11pt]{article}

\usepackage[T1]{fontenc}
\usepackage[utf8]{inputenc}
\usepackage{lmodern}
\usepackage[margin=1in]{geometry}
\usepackage{setspace}
\usepackage[mathlines]{lineno}
\usepackage{booktabs}
\usepackage{array}
\usepackage{multirow}
\usepackage{graphicx}
\usepackage[table]{xcolor}
\usepackage{tikz}
\usetikzlibrary{arrows.meta,positioning,fit,calc}
\usepackage[numbers,sort&compress]{natbib}
\usepackage[font=small,labelfont=bf]{caption}
\usepackage{enumitem}
\usepackage{xurl}
\usepackage[hidelinks]{hyperref}

\newcommand{\pp}{\,pp}          % percentage points
\newcommand{\alttext}[1]{}      % figure alt text: in the Word file only (make_docx.py)
\newcommand{\wordcount}[1]{}    % title-page word count: in the Word file only (make_docx.py)

\title{A local language model layered with rule-based and neural detectors for clinical text de-identification: an open-source toolkit and evaluation}

\author{Valentin Burkin\\[2pt]
\small Independent Researcher, Mason, Ohio, United States\\
\small ORCID \href{https://orcid.org/0009-0001-9497-9496}{0009-0001-9497-9496}\\
\small Correspondence: \href{mailto:valburkin@gmail.com}{valburkin@gmail.com}\wordcount{Word count (main text): 3,987; 4 tables, 2 figures}}

\date{}

\begin{document}

\maketitle

% ---------------------------------------------------------------------------
\begin{abstract}
\noindent\textbf{Objectives:} To measure whether a locally hosted large language model (LLM), stacked with deterministic and clinical named-entity recognition (NER) detectors, finds more protected health information (PHI) than open tools, and at what precision.

\noindent\textbf{Materials and Methods:} DeidSharp, an open-source .NET toolkit, combines the findings of pluggable detectors (patterns, a clinical NER model, a local 26B mixture-of-experts LLM) and fails closed on any detector error. Configurations fixed on tuning halves were compared, under a prespecified protocol, with deid, Philter, the Stanford/Penn de-identifier, OpenAI Privacy Filter and GLiNER on reporting halves of PhysioNet's nursing notes (889 spans) and semi-synthetic MIMIC-IV-Note discharge summaries (29,023 spans), and on synthetic ASQ-PHI queries, with paired patient-level bootstrap intervals.

\noindent\textbf{Results:} In the nursing notes, against Stanford/Penn, the recall-first stack found 98.9\% of PHI (+2.8, 95\% CI 1.4 to 4.4) at 9.3 points lower precision; the precision-first stack, without NER, found 96.5\%, matching Stanford/Penn, at 19.6 points higher precision (83.2\%). In the discharge summaries, both stacks led Stanford/Penn by more than 20 points (recall-first 95.3\%, +21.5, 20.5 to 22.4), almost entirely on generated untyped names; on typed labels all three exceeded 99\%. All four prespecified comparisons held. No configuration beat deid significantly on its home corpus; Philter found more discharge-summary spans than the recall-first stack at 23 points lower precision.

\noindent\textbf{Discussion:} The LLM was the main detector and NER added place names; asked to verify its own findings, the LLM discarded a quarter of true PHI.

\noindent\textbf{Conclusion:} A local LLM layered with deterministic detectors offers recall-first and precision-first operating points without sending text off-site.
\end{abstract}


% ---------------------------------------------------------------------------
\section*{Lay summary}
Doctors' and nurses' notes contain names, dates, places and numbers that identify patients. Before notes can be shared for research, those details must be removed, and doing it by hand does not scale. This work presents a free, open-source tool that combines several automatic methods: fixed rules for numbers and dates, a model trained on clinical notes, and a general-purpose language model that runs on one graphics card inside the hospital, so no text leaves the site. If any part of the tool fails, it returns nothing rather than a partly cleaned note. The test plan was fixed in advance, and the tool was tested on two collections of hospital notes whose identifiers had been replaced with realistic fakes, and on a public set of synthetic queries. With the language model, the tool found 94\% to 99\% of the identifying details: as many as or more than an openly available clinical model, and as many as an older rule-based tool on that tool's own notes. Finding more identifiers usually meant removing more ordinary text as well. All code and results are public, so others can repeat the measurements.

% ---------------------------------------------------------------------------
\section{Background and significance}

Free-text clinical notes hold much of the information in an electronic health record, and sharing them for research or for training models requires removing the identifiers that HIPAA's Safe Harbor method lists, or an equivalent standard \citep{cfr45_164514}. Manual de-identification does not scale, so automated systems have been studied for two decades: rule- and dictionary-based tools such as deid \citep{neamatullah2008} and Philter \citep{norgeot2020}, shared tasks that standardized evaluation \citep{uzuner2007,stubbs2015i2b2,stubbs2015annotating}, and neural models trained on annotated notes \citep{dernoncourt2017,johnson2020chil,chambon2023}. Surveys note that results depend strongly on the corpus and on how matches are counted \citep{meystre2010,aghakasiri2025survey}.

Two problems persist. First, trained models generalize unevenly: one transformer trained on i2b2 found 39\% of the names in PhysioNet's nursing notes, where dictionary tools found 97\% \citep{moore2023pyclipse}, and authors of trained de-identifiers advise local validation before use \citep{chambon2023}. Second, the strongest general language models are hosted services, and sending notes to them is often barred by data use agreements and institutional policy \citep{phys_llm_policy}.

Large language models (LLMs) have been applied to de-identification through hosted APIs \citep{liu2023deidgpt,redactor2025,palacios2026,altalla2025,kocaman2025}, as local models used directly \citep{wiest2025} or fine-tuned on annotated or synthetic notes \citep{doremus2025,lppa2025}, as teachers for smaller models \citep{shield2026}, and in an open on-premises engine with keyed surrogates and an LLM-distilled recognizer \citep{tide2026}. Closest to this work, Zhou et al.\ paired the RoBERTa model used here with a local LLM that verifies and expands its spans \citep{zhou2026chase}, and MedDeID compared open de-identifiers, including a local LLM, with one scorer on the synthetic ASQ-PHI benchmark \citep{hellemans2026meddeid,weatherhead2026asqphi}. Gold standards themselves miss PHI \citep{palacios2026}, which has prompted evaluation without gold labels \citep{teamphi2025}. Few studies compare rule-based tools, trained models and LLM layers with one scorer on the same held-out clinical notes, with paired uncertainty and claims stated before measuring, or report the configurations that failed.

% ---------------------------------------------------------------------------
\section{Objective}

This study asks whether a locally hosted LLM, stacked with deterministic detectors and a clinical NER model, finds more PHI than open tools, and what it costs in precision and in clinical text removed. Its contributions are an open-source, fail-closed toolkit built around that layering; an evaluation against five open baselines with one scorer and paired intervals, under a protocol specified before the final run; a semi-synthetic discharge-summary corpus any credentialed user can rebuild from MIMIC-IV-Note; and a log of every development measurement, including failed designs.

% ---------------------------------------------------------------------------
\section{Materials and methods}

\subsection{The toolkit}

\begin{figure}[t]
\centering
\begin{tikzpicture}[
  font=\footnotesize,
  box/.style={draw, rounded corners=2pt, align=center, inner sep=2pt},
  det/.style={box, fill=blue!6, minimum width=35mm, minimum height=9.5mm},
  stage/.style={box, fill=gray!8, minimum width=28mm, minimum height=12.5mm},
  opt/.style={stage, dashed, fill=orange!8},
  arr/.style={-{Stealth[length=2mm]}, thick}]
  % Detectors: one size, evenly spaced, centered on y = 0.
  \node[det] (k) at (0, 23mm) {Known identifiers};
  \node[det] (p) at (0, 11.5mm) {Patterns};
  \node[det, dashed] (g) at (0, 0) {Gazetteer\\(only without an LLM)};
  \node[det] (c) at (0, -11.5mm) {Clinical NER sidecar\\(Presidio + RoBERTa)};
  \node[det] (l) at (0, -23mm) {Local LLM\\(Ollama, JSON schema)};
  \node[draw, dotted, fit=(k)(l), inner sep=2.5mm, label={[font=\footnotesize]above:Detector stack (parallel; any failure stops the run)}] (stack) {};
  % Input on the stack's center line; the pipeline in one column, then the output row.
  \node[stage, minimum width=27mm, anchor=east] (in) at ($(stack.west)+(-6mm, 0)$) {Note\\(+ record's known\\identifiers)};
  \node[stage, anchor=west] (u) at ($(stack.east)+(6mm, 17.5mm)$) {Union of spans\\+ Safe Harbor\\age filter};
  \node[opt] (so) at ($(u)+(0, -17.5mm)$) {Optional LLM\\second opinion\\(veto NER spans)};
  \node[stage] (r) at ($(so)+(0, -17.5mm)$) {Merge spans;\\replace with tags\\or surrogates};
  \node[stage, fill=green!8, minimum width=22mm, anchor=west] (lg) at ($(r.east)+(5mm, 0)$) {Deterministic\\leak guard};
  \node[stage, minimum width=21mm, anchor=west] (out) at ($(lg.east)+(5mm, 0)$) {De-identified\\text};
  \node[above=2mm of lg, align=center, font=\scriptsize, text=red!60!black] {refuses output with\\a surviving identifier};
  \draw[arr] (in.east) -- (stack.west);
  \draw[arr] (stack.east |- u.west) -- (u.west);
  \draw[arr] (u) -- (so);
  \draw[arr] (so) -- (r);
  \draw[arr] (r) -- (lg);
  \draw[arr] (lg) -- (out);
\end{tikzpicture}
\caption{DeidSharp's free-text pipeline. Detectors return positions and categories only. The stack replaces every span any detector reports; the optional second opinion may drop only spans from the clinical NER model that are not locations or organizations. The leak guard uses the deterministic detectors only and was not exercised in this evaluation.}
\label{fig:architecture}

\alttext{A flow diagram. A note, with any identifiers the patient record already holds, enters a stack of five detectors run in parallel: known identifiers, patterns, a gazetteer used only without a language model, a clinical NER sidecar, and a local language model. Their spans are unioned and filtered for ages under 90, optionally checked by a language-model second opinion, merged and replaced, and the result passes a deterministic leak guard before output.}
\end{figure}
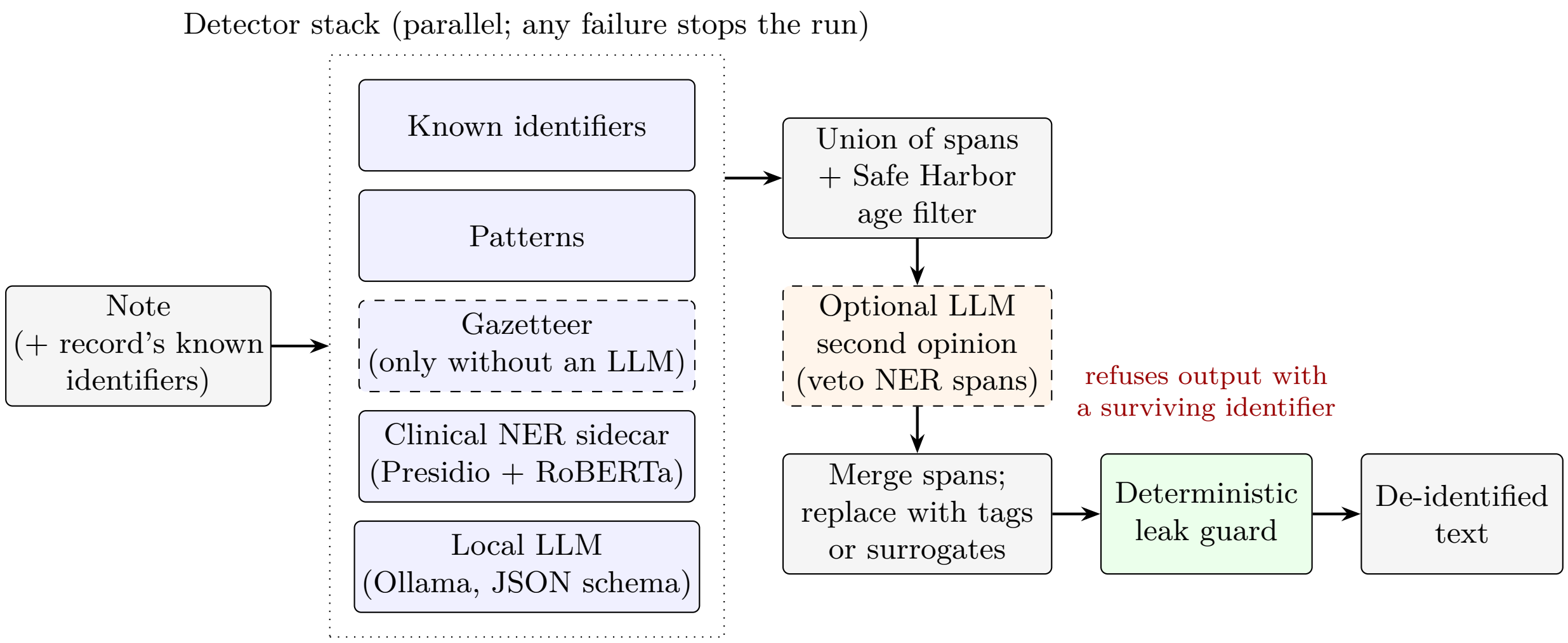

DeidSharp is a set of MIT-licensed .NET 10 libraries \citep{deidsharp2026}. Besides Safe Harbor transforms for structured data and FHIR R4 resources \citep{hl7fhir} and keyed pseudonyms, it de-identifies free text with a stack of detectors run in parallel (Figure~\ref{fig:architecture}). Detectors return spans and categories, never the covered text, so logs and errors never quote PHI. The union of all spans is replaced (recall first); if any detector fails, the run fails rather than reading the failure as ``nothing found.'' A deterministic leak guard then refuses output in which a known identifier survived. This study evaluates detection only: the corpora supply no structured record, so neither the known-identifier detector nor the leak guard was exercised.

\paragraph{Detectors.} The \emph{known-identifier} detector matches the names, dates and numbers the structured record already holds (none here, so this study's numbers are a floor for use with a record). The \emph{pattern} detector holds rules for shaped identifiers (dates, phones, record numbers, emails, URLs and others, with checksums where they exist). The \emph{gazetteer} matches US Census name and place lists, with stop lists to limit false matches. The \emph{clinical NER} detector calls a Presidio analyzer \citep{presidio} running the obi/deid\_roberta\_i2b2 RoBERTa model, trained on the i2b2 2014 corpus \citep{obi_roberta}, in a local container. Presidio's general spaCy NER and its URL and US-license recognizers were disabled: on the nursing notes' tuning half they produced most of the false positives for about 2 points of recall (Results).

\paragraph{The LLM detector.} The LLM detector sends text in chunks of 2,000 characters overlapping by 200 to a local model through Microsoft.Extensions.AI's chat-client abstraction, here served by Ollama \citep{ollama}, at temperature 0 with thinking disabled and a JSON schema that constrains the answer to a list of \{text, type\} items. Each listed text is located in the note wherever it occurs as whole words; text the model invented or reworded is dropped. An answer cut off at the output limit is requested again in halves and fails closed if it still cannot fit. Rule filters drop what Safe Harbor permits keeping, whatever type the model assigned: ages under 90, a year, time or weekday alone, and identifiers of three digits or fewer. The same age rule wraps every stack, since NER models also tag every age. The prompt holds instructions and forty short fictional notes with their answers (about 5,000 tokens; twelve notes contain no PHI); an AI assistant wrote them from fictional examples of the tuning half's misses, never from corpus text; the model's context window was raised to 8,192 tokens so the prompt is never truncated. The model of record is Gemma~4 26B-A4B \citep{gemma4}, a mixture of experts with about 4B active parameters, in Ollama's default 4-bit quantization. The prompts are in the repository.

\paragraph{Second opinion.} An optional check shows the LLM each candidate span in its note (with a 16,384-token context window, to fit long notes) and drops the ones it names as not PHI; it is told to keep a span when unsure. It may veto only spans no known-identifier, pattern or LLM detector reported, and never a location or organization (Results).

\paragraph{Stacks.} Five configurations were fixed on the nursing notes' tuning half: \emph{recall first} (known identifiers, patterns, clinical NER, LLM); \emph{balanced} (the same with the second opinion); \emph{balanced, one call} (known identifiers, patterns, only the locations and organizations of the clinical NER model, LLM); \emph{precision first} (known identifiers, patterns, LLM; no NER sidecar); and \emph{no LLM} (known identifiers, patterns, gazetteer, clinical NER). Stacks with the LLM omit the gazetteer (Results).

\subsection{Corpora}

Table~\ref{tab:corpora} summarizes the corpora. The \emph{nursing notes} are PhysioNet's Deidentified Medical Text: 2,434 ICU nursing notes from 163 patients, whose PHI was replaced with realistic surrogates, with the gold standard distributed with the deid software \citep{douglass2007physionet,neamatullah2008,deid2008software}. They were deid's development set, so deid's default configuration includes lists built from them. The \emph{discharge summaries} are a semi-synthetic corpus constructed for this study from MIMIC-IV-Note 2.2 \citep{johnson2023mimicivnote,johnson2023mimiciv}, whose de-identification replaced each PHI item with ``\_\_\_'' and kept no annotation \citep{johnson2020chil}. Following the ``hiding in plain sight'' approach \citep{carrell2013}, context rules decide what each placeholder stood for (``Dr.~\_\_\_'' a clinician's surname, ``\_\_\_ 07:15AM'' a date), and a fictional identity per note fills it; each surrogate becomes a gold span. The gold standard is therefore what MIMIC's own de-identification found. Placeholders that match no rule (41\% on the tuning half) are filled with a first name or surname, after a random sample of 100, read in context, showed that every one stood for a name; these are reported as \emph{untyped names}. The corpus (1,000 patients, one summary each) is drawn by a fixed hash, so any credentialed user rebuilds the identical corpus. \emph{ASQ-PHI} is a public synthetic benchmark of 1,051 clinical search queries written by GPT-4o, 219 of them hard negatives without PHI \citep{weatherhead2026asqphi}; its tags carry values, not offsets, so every occurrence of a tagged value in its query is a gold span.

\begin{table}[!htbp]
\centering
\small
\caption{Corpora. Spans are gold spans in the Safe Harbor scope. ASQ-PHI has no held-out half: its counts are the whole corpus, which was consulted throughout development.}
\label{tab:corpora}
\begin{tabular}{@{}l>{\raggedright\arraybackslash}p{5.6cm}rrrr@{}}
\toprule
& & \multicolumn{2}{c}{Tuning half} & \multicolumn{2}{c}{Reporting half} \\
\cmidrule(lr){3-4}\cmidrule(l){5-6}
Corpus & Content & Docs & Spans & Docs & Spans \\
\midrule
Nursing notes & ICU nursing notes, surrogate PHI, gold from deid's authors (PhysioNet, credentialed) & 1,127 & 844 & 1,307 & 889 \\
Discharge summaries & MIMIC-IV-Note 2.2, placeholders refilled with generated surrogates (credentialed) & 506 & 30,054 & 494 & 29,023 \\
ASQ-PHI & Synthetic clinical search queries (MIT license) & \multicolumn{2}{c}{---} & 1,051 & 2,976 \\
\bottomrule
\end{tabular}

\end{table}

\subsection{Development and use of the reporting halves}

Each credentialed corpus is split by patient into a \emph{tuning} half, on which misses and false positives were read and detectors changed, and a \emph{reporting} half for published numbers; a patient's notes fall on one side by a fixed hash of the patient's number. Detector settings were fixed on the nursing notes' tuning half; the discharge summaries' tuning half was used to build that corpus and for later checks of the prompt and model family. Every change was logged as a dated finding with its command, commit and counts, including failed ideas; the log is public, and the commits it names before release 1.0.0 belong to the pre-release development history (\texttt{REPRODUCING.md} in the repository). Both reporting halves were measured before the final run. The nursing notes' reporting half was measured five times during development, each time for configurations already fixed on the tuning half, though one of those measurements (the recall the clinical sidecar lost without spaCy's NER) also informed keeping the gazetteer in the stack without an LLM. The discharge summaries' reporting half was measured once before the final run; during that measurement the second opinion's context window was raised to fit long notes, after the results without an LLM had been read and before any result with one. Both halves were measured again for the final run and its sensitivity check, and no setting was changed after a final-run result was read.

The toolkit's code, tests, documentation and prompt examples were written with an AI coding assistant under the author's direction (see Use of generative AI). No note text from either credentialed corpus was given to the assistant: it worked only through the evaluation tool, whose output is counts and rates, and misses were read locally, outside the assistant, and described to it with fictional examples. During the study this rested on the assistant's instructions and the evaluation tool's counts-only output; a later audit of the assistant's session logs found no note text, and access to the data is also blocked in its configuration. Every model that processed corpus text, the LLM included, ran on one local workstation.

\subsection{Baselines}

Five open tools were run in their default or published configurations, with nothing tuned on these corpora, and scored from files of their predicted spans. \emph{deid} 1.1 \citep{neamatullah2008} ran in three configurations: its default, which uses corpus-specific lists built for the nursing notes (surrogate patient names, staff and local-place lists, a ward rule); without those lists; and without them and its hospital list. \emph{Philter} (philter-ucsf 1.0.3) ran in its default configuration \citep{norgeot2020}. Three models ran from weights pinned by commit in an offline container on the GPU: the \emph{Stanford/Penn} de-identifier \citep{chambon2023} as a standard token-classification pipeline with simple span aggregation (chosen for this study), without the rule-based ``hide in plain sight'' component its authors combine it with; \emph{OpenAI Privacy Filter} \citep{openai2026privacyfilter} through its own package with constrained decoding at the default operating point; and \emph{GLiNER multi-PII} \citep{zaratiana2024gliner} at threshold 0.5, asked for the Safe Harbor identifier types in plain words chosen for this study, in windows of 160 words overlapping by 40. To check the scorer, deid's default output on all 2,434 nursing notes, scored with every label, reproduced its published results exactly (recall 96.68\%, 59 spans missed, precision 74.83\%; published: 0.967, 59 and 0.749 \citep{neamatullah2008}).

\subsection{Measures and statistics}

The headline measure is \emph{span recall}: the share of gold spans that any prediction overlaps, whatever category it assigns, since a gold span no prediction touches is leaked PHI. \emph{Precision} is the share of predicted spans that overlap a gold span. Also reported are \emph{character recall} (gold characters covered, as in \citep{hellemans2026meddeid}), the share of \emph{non-PHI text removed} (non-whitespace characters outside gold that a prediction covers), word-level F1, and an exact-boundary \emph{strict} match comparable to the binary evaluation of the 2014 i2b2/UTHealth task \citep{stubbs2015i2b2} (Supplementary Table S1). The scope is Safe Harbor: a bare year and ages under 90 are not PHI, and a prediction touching only such gold counts as a false positive (a sensitivity analysis counting them neither way is in Supplementary Table S2). Intervals are percentile bootstraps over 1,000 resamples of patients (queries for ASQ-PHI) with a fixed seed \citep{efron1993}; a patient's notes share names and dates, so they are resampled together. The nursing notes' 163 patients are split between the halves, so their intervals are wide; each discharge summary comes from a different patient. Every stack and baseline in a run is measured on the same resamples, and comparisons use the paired difference's interval.

\subsection{Prespecified final run}

A protocol fixing the final run's data, stacks, baselines, measures and four statements was committed to the project repository before any run on a reporting half for that run; the commit (Git tag \texttt{d6-protocol}) became public together with the results, so its timing rests on the project's own Git log, not on an external registry. When it was written, the stacks' earlier reporting-half results and the baselines' tuning-half results were known, so the statements, especially on the discharge summaries, guarded against a regression more than they tested a new effect. Comparisons with deid, Philter and the other models were not prespecified and are exploratory. The statements compare the stacks with the Stanford/Penn de-identifier, chosen as the model baseline with the highest span recall on both tuning halves and on ASQ-PHI. On each reporting half, \emph{statement A}: the recall-first stack finds more PHI (paired recall interval above zero); \emph{statement B}: the precision-first stack loses at most 2 points of recall (paired interval's lower bound above $-2$, a margin chosen without formal derivation) and has higher precision (paired interval above zero). The four statements are each read from their own 95\% interval, without adjustment for multiplicity. Nothing was stated beforehand for ASQ-PHI, where the stacks and the Stanford/Penn model were already near the ceiling in development. One amendment, made after a first pass on ASQ-PHI and the nursing notes had been read, moved the model baselines to a GPU build of the same image and reran every corpus on one commit; the superseded pass gave the same verdicts. Each corpus was then run once, on commit \texttt{2059384} (the libraries at version 1.1.0; the archived 1.1.0 release adds publishing safeguards and the sensitivity-check option to the evaluation tool but changes no detector), with the LLM pinned by its Ollama digest and server version (0.40.11), on one workstation (NVIDIA RTX 5080 with 16~GB, Intel Core Ultra 9 285K, 64~GB of memory) with the clinical sidecar on the CPU.

% ---------------------------------------------------------------------------
\section{Results}

\subsection{Development on the tuning halves}

Table~\ref{tab:development} summarizes the development findings that shaped the stacks. Within Gemma, recall and precision rose with total model size; within Qwen, a 35B mixture of experts with 3B active parameters gained precision but lost recall; Gemma~4 26B-A4B was best on every measure, at about 0.4\,s per note. In an ablation, the LLM alone found 95.6\% of PHI at 87.6\% precision, the clinical NER model alone 87.0\% at 58.3\%, and their union 98.7\%: the NER model's contribution was mostly locations (it recovered 19 of the 23 places the LLM missed), at a large cost in precision. Pattern detectors added no measurable recall next to the LLM; they remain in every stack because they are deterministic, feed the leak guard, and match the record's own identifiers. All forty fictional examples gave 10 points more precision than four, at the same speed, and beat examples retrieved per note. A zero-shot prompt found as much PHI as the examples with Gemma on the nursing notes (35 spans missed each) and 0.8 points more on the discharge summaries, at lower precision, so the examples bought precision, not recall. A third family (Ministral~3 14B \citep{ministral3}) gained as much recall over the deterministic layer (+70.9 points zero-shot against Gemma's +70.1) at far lower precision.

The second opinion raised precision by 18.7 points when allowed to judge every span from the clinical model, but rejected real hospital and unit names; barred from locations and organizations, it cost 2 spans ($-0.24$\pp, $-0.57$ to 0.00) for +11.9 points of precision. Allowed to judge the LLM's own findings as well, the same model discarded 24.6 points of recall ($-29.5$ to $-19.8$) for +3.3 points of precision.

\begin{table}[!htbp]
\centering
\small
\caption{Selected development findings on the nursing notes' tuning half (844 gold spans). Recall is span recall, with its 95\% interval where it decided a choice. ``Det.'' is the known-identifier and pattern detectors; LLM is gemma4:26b-a4b unless named.}
\label{tab:development}
\begin{tabular}{@{}p{8.4cm}rr@{}}
\toprule
Configuration & Recall (\%) & Precision (\%) \\
\midrule
\multicolumn{3}{@{}l}{\textit{Clinical NER sidecar}} \\
\quad Det. + clinical sidecar, every Presidio recognizer & 89.7 (86.4--92.5) & 4.9 \\
\quad \ldots without spaCy NER, URL and license recognizers & 87.7 (84.1--90.5) & 55.0 \\
\multicolumn{3}{@{}l}{\textit{Model size and family (det. + gazetteer + LLM, four examples)}} \\
\quad gemma4:e4b / 12b / 26b-a4b & 88.6 / 93.4 / 96.8 & 40.0 / 50.2 / 52.6 \\
\quad qwen3.5:9b / 35b-a3b & 93.8 / 91.6 & 28.5 / 45.7 \\
\multicolumn{3}{@{}l}{\textit{Gazetteer next to the LLM}} \\
\quad Det. + LLM with / without gazetteer & 96.7 / 96.3 & 51.7 / 71.7 \\
\multicolumn{3}{@{}l}{\textit{Prompt (det. + LLM)}} \\
\quad Four fixed / four retrieved / all forty examples & 96.2 / 96.3 / 95.7 & 74.1 / 77.0 / 84.2 \\
\quad Zero-shot / forty examples (another run) & 95.9 / 95.9 & 76.9 / 83.8 \\
\multicolumn{3}{@{}l}{\textit{Another family (det. + ministral-3:14b)}} \\
\quad Zero-shot / forty examples & 96.6 / 93.4 & 12.9 / 38.2 \\
\multicolumn{3}{@{}l}{\textit{Ablation (forty examples, no gazetteer)}} \\
\quad Patterns / clinical NER / LLM alone & 25.7 / 87.0 / 95.6 & 87.1 / 58.3 / 87.6 \\
\quad Clinical NER + LLM / with patterns & 98.7 / 98.7 & 57.1 / 56.2 \\
\quad Clinical places only + LLM & 97.9 (96.5--98.9) & 74.7 \\
\multicolumn{3}{@{}l}{\textit{Second opinion on det. + clinical + LLM (four examples)}} \\
\quad None / on every category / barred from places & 99.1 / 97.9 / 98.8 & 50.8 / 69.5 / 62.7 \\
\multicolumn{3}{@{}l}{\textit{Self-check (det. + LLM, forty examples)}} \\
\quad Second opinion on the LLM's own spans & 71.1 (65.8--76.4) & 87.5 \\
\bottomrule
\end{tabular}
\end{table}

\subsection{Final run}

\begin{figure}[!htb]
\centering
\includegraphics[width=\textwidth]{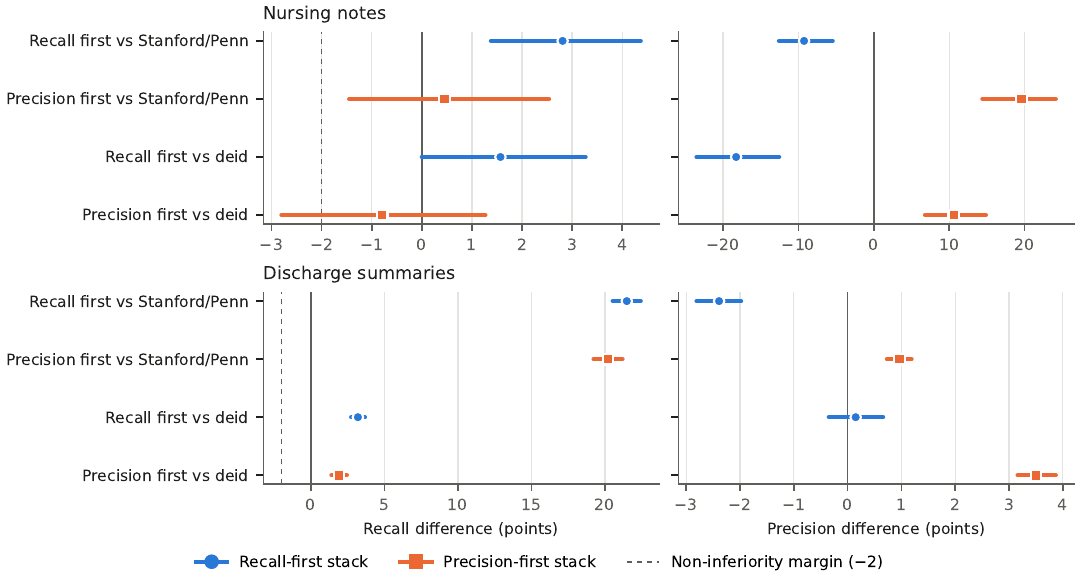}
\caption{Paired differences in span recall (left) and precision (right), in percentage points with paired 95\% bootstrap intervals, between the two prespecified stacks and the Stanford/Penn de-identifier or deid with its corpus-specific lists, on the reporting halves. The dashed line is the non-inferiority margin, which applies to the precision-first stack against Stanford/Penn. Scales differ between panels. Philter, which found more discharge-summary spans than both stacks (paired recall $-4.1$ and $-5.4$ points) at 23 to 27 points lower precision, is compared in the text. Supplementary Figure S1 shows every stack and tool on an absolute scale.}
\label{fig:differences}

\alttext{Four panels of horizontal interval plots. For the nursing notes, the recall-first stack has higher recall than Stanford/Penn (interval above zero) and lower precision; the precision-first stack's recall interval crosses zero but stays above the minus-2 margin, with precision about 20 points higher. Against deid, both stacks' recall intervals include zero. For the discharge summaries, both stacks lead Stanford/Penn by about 20 points of recall and deid by 2 to 3 points.}
\end{figure}

Table~\ref{tab:results} gives the final run, Figure~\ref{fig:differences} its paired comparisons, and Table~\ref{tab:statements} the prespecified statements, all of which held. On the nursing notes, the recall-first stack missed 10 of 889 gold spans, against 35 for Stanford/Penn and 24 for deid with its corpus-specific lists, and paid 9.3 points of precision relative to Stanford/Penn ($-12.6$ to $-5.5$). The precision-first stack, with no NER sidecar, matched Stanford/Penn's recall (+0.45, $-1.45$ to 2.54) at 19.6 points more precision. On the discharge summaries, both stacks led Stanford/Penn by more than 20 points of recall. That margin came from the untyped names (12,632 spans, 44\% of gold), of which Stanford/Penn found 40.7\% and the recall-first and precision-first stacks 89.2\% and 86.6\%; on the typed labels, they found 99.3\%, 99.96\% and 99.7\%. The LLM alone, an ablation measured in earlier runs, is in Supplementary Table S5.

\begin{table}[p]
\centering
\scriptsize
\caption{Final run on the reporting halves and ASQ-PHI (Safe Harbor scope), with 95\% bootstrap intervals over patients (queries for ASQ-PHI). ``Text'' is the share of non-PHI text removed. Every stack uses the known-identifier and pattern detectors; LLM: gemma4:26b-a4b with forty fictional examples. ``Lists'': deid's corpus-specific lists for the nursing notes. Gold spans: 889 (nursing notes), 29,023 (discharge summaries), 2,976 (ASQ-PHI).}
\label{tab:results}
\setlength{\tabcolsep}{2.5pt}
\begin{tabular}{@{}llrrrrr@{}}
\toprule
 & Stack or tool & Recall (95\% CI) & Missed & Char. recall & Precision (95\% CI) & Text \\
\midrule
\multirow{11}{*}{\rotatebox{90}{Nursing notes}}
 & Recall first: NER + LLM & 98.88 (97.91--99.58) & 10 & 99.00 & 54.31 (49.45--59.60) & 0.57 \\
 & Balanced: + second opinion & 98.20 (96.86--99.24) & 16 & 98.84 & 70.53 (66.77--74.42) & 0.26 \\
 & One call: NER places + LLM & 97.53 (95.95--98.74) & 22 & 98.29 & 72.03 (68.33--75.80) & 0.24 \\
 & Precision first: LLM & 96.51 (94.92--97.93) & 31 & 97.59 & 83.18 (79.52--86.49) & 0.15 \\
 & No LLM: gazetteer + clinical NER & 88.30 (84.71--91.11) & 104 & 86.38 & 43.02 (38.53--47.64) & 0.85 \\
 \cmidrule(l){2-7}
 & deid, with lists & 97.30 (95.73--98.61) & 24 & 97.73 & 72.57 (67.77--76.77) & 0.18 \\
 & deid, without lists & 96.06 (94.16--97.57) & 35 & 96.41 & 72.39 (67.40--76.72) & 0.17 \\
 & Stanford/Penn & 96.06 (94.56--97.27) & 35 & 96.29 & 63.56 (59.66--67.48) & 0.33 \\
 & Philter & 91.45 (88.99--93.71) & 76 & 88.00 & 6.71 (5.77--7.63) & 7.81 \\
 & GLiNER multi-PII & 76.83 (71.35--81.31) & 206 & 76.61 & 11.08 (9.73--12.67) & 2.48 \\
 & OpenAI Privacy Filter & 32.85 (26.16--38.80) & 597 & 36.35 & 59.16 (52.06--66.39) & 0.18 \\
\midrule
\multirow{11}{*}{\rotatebox{90}{Discharge summaries}}
 & Recall first & 95.28 (94.95--95.59) & 1,371 & 97.08 & 95.94 (95.61--96.27) & 0.78 \\
 & Balanced & 94.91 (94.59--95.23) & 1,476 & 96.86 & 96.70 (96.39--97.02) & 0.74 \\
 & Balanced, one call & 94.55 (94.18--94.91) & 1,582 & 96.61 & 96.82 (96.51--97.12) & 0.70 \\
 & Precision first & 94.00 (93.63--94.38) & 1,742 & 96.00 & 99.29 (99.17--99.41) & 0.61 \\
 & No LLM & 91.57 (91.17--91.98) & 2,446 & 94.21 & 94.76 (94.40--95.13) & 0.42 \\
 \cmidrule(l){2-7}
 & deid, with lists & 92.08 (91.67--92.47) & 2,299 & 84.32 & 95.78 (95.43--96.13) & 0.28 \\
 & deid, without lists & 91.73 (91.30--92.12) & 2,399 & 84.13 & 95.81 (95.45--96.14) & 0.28 \\
 & Stanford/Penn & 73.79 (72.82--74.74) & 7,608 & 82.11 & 98.33 (98.10--98.56) & 0.25 \\
 & Philter & 99.41 (99.31--99.52) & 171 & 83.34 & 72.45 (71.84--73.10) & 2.45 \\
 & GLiNER multi-PII & 61.70 (60.83--62.54) & 11,115 & 66.53 & 56.72 (55.47--58.14) & 2.53 \\
 & OpenAI Privacy Filter & 28.80 (27.71--29.85) & 20,664 & 29.78 & 96.36 (95.89--96.88) & 0.24 \\
\midrule
\multirow{10}{*}{\rotatebox{90}{ASQ-PHI}}
 & Recall first & 99.73 (99.53--99.90) & 8 & 97.87 & 96.65 (95.92--97.28) & 0.57 \\
 & Balanced & 99.66 (99.46--99.86) & 10 & 97.83 & 99.12 (98.73--99.43) & 0.24 \\
 & Balanced, one call & 99.66 (99.46--99.86) & 10 & 97.79 & 99.12 (98.73--99.43) & 0.22 \\
 & Precision first & 99.63 (99.40--99.83) & 11 & 97.68 & 99.57 (99.33--99.78) & 0.14 \\
 & No LLM & 99.73 (99.53--99.90) & 8 & 96.38 & 95.30 (94.45--96.08) & 0.86 \\
 \cmidrule(l){2-7}
 & deid, with lists & 85.85 (84.82--86.98) & 421 & 68.45 & 94.69 (93.76--95.50) & 1.52 \\
 & Stanford/Penn & 99.66 (99.45--99.87) & 10 & 98.23 & 97.71 (97.15--98.20) & 0.56 \\
 & Philter & 96.74 (96.13--97.34) & 97 & 70.68 & 93.55 (92.80--94.30) & 2.18 \\
 & GLiNER multi-PII & 99.23 (98.90--99.52) & 23 & 99.00 & 70.74 (69.41--72.02) & 13.26 \\
 & OpenAI Privacy Filter & 70.19 (68.95--71.39) & 887 & 64.57 & 96.66 (95.83--97.42) & 1.60 \\
\bottomrule
\end{tabular}
\end{table}

\begin{table}[!htbp]
\centering
\small
\caption{Prespecified statements, each a paired comparison with the Stanford/Penn de-identifier (difference in percentage points, paired 95\% bootstrap interval). Statement B has two conditions.}
\label{tab:statements}
\begin{tabular}{@{}lllrl@{}}
\toprule
Reporting half & & Condition & Paired difference & Held \\
\midrule
Nursing & A & Recall first finds more PHI & recall +2.81 (1.38 to 4.37) & Yes \\
 & B & Precision first loses $<2$ points recall & recall +0.45 ($-1.45$ to 2.54) & \multirow{2}{*}{Yes} \\
 & & \ldots and has higher precision & precision +19.62 (14.40 to 24.14) & \\
Discharge & A & Recall first finds more PHI & recall +21.49 (20.53 to 22.44) & Yes \\
 & B & Precision first loses $<2$ points recall & recall +20.21 (19.23 to 21.20) & \multirow{2}{*}{Yes} \\
 & & \ldots and has higher precision & precision +0.96 (0.73 to 1.19) & \\
\bottomrule
\end{tabular}
\end{table}

\subsection{Exploratory comparisons}

On ASQ-PHI, no stack with the LLM differed significantly from Stanford/Penn in recall (recall-first +0.07, 0.00 to 0.17), and the precision-first stack had 1.86 points more precision (1.39 to 2.39). On the nursing notes, deid's home corpus, no configuration found significantly more PHI than deid with its corpus-specific lists: the recall-first stack's lead (+1.57 points) had a paired interval starting at zero (0.00 to 3.27), and the precision-first stack's recall did not differ significantly ($-0.79$, $-2.80$ to 1.27) at 10.6 points more precision (6.8 to 14.9). Against deid without those lists, the recall-first and balanced stacks found more (+2.81, 1.04 to 4.80; +2.14, 0.33 to 4.04). Without an LLM, the toolkit trailed deid by 9 points of recall. On the nursing notes, whose gold standard follows deid's conventions, deid also had the highest strict (exact-boundary) F1 (Supplementary Table S1). On the discharge summaries, every stack with the LLM found more than deid in any configuration (recall-first +3.20, 2.74 to 3.68, at equal precision; precision-first +1.92, 1.40 to 2.45, with 3.5 points more precision). Much of that lead is the generator's fictional hospitals and cities, which deid's lists rarely match; without them, the leads shrink to about 1.5 and 0.2 points (unpaired). Philter touched more gold spans on the discharge summaries (99.4\%) but covered fewer PHI characters (83.3\% against 96.0--97.1\%), removed three to four times as much other text, and had 23 to 27 points less precision.

\subsection{Reproduction, sensitivity and runtime}

The final run reproduced earlier measurements made at other commits: exactly on the discharge summaries, to within 0.01 point on ASQ-PHI, and to within one span on the nursing notes. Three consecutive runs had given identical spans on the nursing notes and ASQ-PHI; the discharge summaries' LLM stacks ran once. Counting out-of-scope gold neither for nor against any tool left recall unchanged, moved precision by at most 1.2 points, and changed no verdict (Supplementary Table S2). The LLM took 0.4\,s per nursing note and 7.5\,s per discharge summary, with a fifth of the 17.9~GB model in CPU memory; the full run took 24 minutes each for the nursing notes and ASQ-PHI and 2 hours 14 minutes for the discharge summaries.

% ---------------------------------------------------------------------------
\section{Discussion}

\subsection{Principal findings}

Under a prespecified protocol, the recall-first stack found more PHI than the strongest open clinical model on both reporting halves, and the precision-first stack, without NER, matched it on the nursing notes and exceeded it on the discharge summaries, with higher precision on both. The discharge-summary margins come from the corpus's generated untyped names; on typed labels Stanford/Penn and both stacks found more than 99\%. In the ablations, the LLM was the main detector; the clinical NER model, trained with i2b2's hospital label, added place names; and patterns added little. Because the two models fail differently, their union beat either one.

Three results bear on how LLMs should be used for de-identification. First, the prompt's fictional examples and the model's family decided precision more than recall: a zero-shot prompt found as much PHI, and a third family gained as much recall over the deterministic layer at far lower precision; within Gemma, a model with more total parameters raised both. Second, a model is a poor judge of its own findings: asked to list PHI, it found it; asked to verify the same spans as candidates, it discarded a quarter of the true PHI. Verification helped only when confined to another detector's spans and barred from places. Third, a Census gazetteer worth about 3 points of recall without an LLM added 2 or 3 spans next to one, at a large cost in precision; local site lists, such as deid's hospital list (11 points on its home corpus), were not tested next to an LLM.

\subsection{Comparison with prior work}

Zhou et al.\ also found that adding an LLM to the same RoBERTa model reduced false negatives, though it lowered entity-level F1 on i2b2; their LLM verified the NER model's spans, the role in which verification helped here too \citep{zhou2026chase}. Larger local models also de-identified better in other work \citep{wiest2025}, and multi-pass and auditor designs did not beat a single pass \citep{palacios2026}. On ASQ-PHI, MedDeID's own model reached a character recall of 98.9\%, above the DeidSharp stacks' 97.7\% to 97.9\%; its comparator scores are not directly comparable with those here, since it post-processed their spans and configured some differently \citep{hellemans2026meddeid}.

These comparisons need qualifying. The baselines ran untuned and out of domain, while DeidSharp's prompt examples were written after the tuning half's misses had been read; Stanford/Penn ran without the rule-based component its authors pair it with; and no published LLM-based de-identifier was a baseline. The rule-based tools held their ground: deid matched every stack on its home corpus, and Philter found more discharge-summary spans at far lower precision. Results also varied by corpus: Stanford/Penn matched the precision-first stack on nursing notes and synthetic queries but found 41\% of untyped names in discharge summaries, and ASQ-PHI, where the stacks, Stanford/Penn and GLiNER all exceeded 99\% recall, separated the tools less than the clinical corpora did.

\subsection{Implications for practice}

Each stack suits a different setting. Without a GPU, the stack without an LLM (its NER sidecar on a CPU) found 88\% of the nursing notes' PHI at 43\% precision. With one 16~GB GPU and no sidecar, the precision-first stack found 96.5\% and removed 0.15\% of the other text. Where recall must come first, the recall-first stack missed 10 of 889 spans at 54\% precision, and the balanced stacks sit between the two. Whichever is chosen, recall should be measured on a sample of the site's own notes.

\subsection{Limitations}

Both credentialed corpora come from one hospital (Beth Israel Deaconess Medical Center) through MIMIC, and both carry surrogate rather than original PHI. The discharge summaries are semi-synthetic: their gold standard is what MIMIC's de-identification found; their surrogates come from rules written for this study, with names from the gazetteer's lists and places on no real list; 44\% of the reporting half's gold spans are untyped names, a label based on a sample of 100 (by the rule of three, up to 3\% may stand for something else); one identity per note and reused clinicians' names favor detectors that find repeated mentions; and generated surrogates may be more regular than real PHI. The nursing notes' gold standard was built partly with deid's prototype, and gold standards miss PHI \citep{palacios2026}; false positives were not reviewed by hand, so some may be PHI the gold missed. ASQ-PHI's queries were generated by an LLM and are public, and the open deid package publishes the nursing notes' PHI terms and deid's scrubbed output of every note, so a model trained on web data may have seen either corpus. The i2b2/n2c2 corpora, on which most prior work reports, were not open to new registration.

The reporting halves were measured during development, and the final run's protocol is self-attested and was written knowing earlier results (Methods). Recall counts any overlap as found, which is lenient next to strict matching (reported in the supplement) and to document-level ``all-or-nothing'' recall \citep{scaiano2016}, which was not reported. A known gazetteer fault, which affects only the stack without an LLM, was left unfixed so that the run measured the code the protocol fixed. The leak guard and the known-identifier detector were not evaluated.

The final run used one LLM and one server version, and another may change spans slightly; only two other model families were tried, on tuning halves. A Stanford/Penn successor that adds ages \citep{prakash2025} was not evaluated, and the tools' overlapping spans, unlike DeidSharp's stacks', were scored unmerged, which slightly affects GLiNER's precision. Finally, the measured stacks need a 16~GB GPU and, for the recall-first stack, a Docker sidecar; no stack is a substitute for local validation and governance.

% ---------------------------------------------------------------------------
\section{Conclusion}

A local LLM layered with deterministic detectors and a clinical NER model gave recall-first and precision-first operating points that, under a prespecified protocol, found more PHI than an open clinical de-identifier on two reporting halves (the precision-first stack matched it on one, with higher precision), largely on names a trained model missed, without sending text off-site. On deid's home corpus they matched rather than surpassed a rule-based tool built for it. The toolkit, the evaluation tool, every finding including negative ones, and the commands to rerun them are public.

% ---------------------------------------------------------------------------
\section*{Author contributions}
V.B.: conceptualization, methodology, software, investigation, formal analysis, data curation, visualization, and writing (review and editing). The software was developed, the evaluation runs executed and the original draft prepared with an AI coding assistant under V.B.'s direction.

\section*{Use of generative AI}
The work used Claude Code (Anthropic) with Claude Opus 5.5, and Claude Haiku 5.5 for minor helper tasks, under the author's direction. The assistant wrote the software, its tests and documentation, and the detector prompts with their fictional examples; it ran the evaluation tool; it proposed experiments, which the author approved or rejected; and it drafted the findings log, the protocols and this manuscript. The author verified its work by reviewing each change, by commissioning independent reviews of the code and of the manuscript's claims from separate instances of the assistant, and by external checks (the scorer reproduced deid's published results exactly). The assistant never received note text from the credentialed corpora (Methods). The author takes full responsibility for the content.

\section*{Funding}
This research received no specific funding.

\section*{Conflicts of interest}
None declared. DeidSharp is non-commercial open-source software released under the MIT license.

\section*{Ethics statement}
This study is a secondary analysis of de-identified data, used under PhysioNet's credentialed data use agreements and within the approvals and consent waivers granted for the source databases by their Institutional Review Boards. The nursing notes come from MIMIC-II; MIMIC-III, which PhysioNet lists as their parent project, was approved by the Institutional Review Boards of Beth Israel Deaconess Medical Center and the Massachusetts Institute of Technology, which waived individual consent \citep{johnson2016mimiciii}; the creation of MIMIC-IV and MIMIC-IV-Note was reviewed by the Institutional Review Board of Beth Israel Deaconess Medical Center, which granted a waiver of informed consent \citep{johnson2023mimiciv,johnson2023mimicivnote}. All processing of note text took place on a local workstation. ASQ-PHI is synthetic and contains no data on real patients.

\section*{Data availability}
The nursing notes \citep{douglass2007physionet} and MIMIC-IV-Note \citep{johnson2023mimicivnote} are available to credentialed users on PhysioNet \citep{goldberger2000physionet,pollard2026physionet}; the discharge-summary corpus is rebuilt deterministically from the MIMIC-IV-Note release by the evaluation tool. ASQ-PHI is public under the MIT license, and a copy ships with the code. Published results (counts and rates) are in \texttt{docs/evaluation} of the repository.

\section*{Code availability}
DeidSharp, the evaluation tool, the baseline containers, the prompts, the findings log and the reproduction instructions are available at \url{https://github.com/valburkin/DeidSharp}. The final run was measured at commit \texttt{2059384}; release 1.1.0, which adds publishing safeguards and the out-of-scope sensitivity option to the evaluation tool, is archived at \url{https://doi.org/10.5281/zenodo.23823966} \citep{deidsharp2026} and published on NuGet. The protocol of the final run is the Git tag \texttt{d6-protocol}.

% ---------------------------------------------------------------------------
\singlespacing
\bibliographystyle{unsrtnat}
\bibliography{references}

\end{document}